\documentclass[runningheads]{llncs}
\usepackage{graphicx}

\usepackage{tikz}
\usepackage{comment}
\usepackage{amsmath,amssymb} 
\usepackage{color}
\usepackage{bm}
\usepackage{cleveref}
\usepackage{multirow}
\usepackage{graphicx}
\usepackage{overpic}
\usepackage{booktabs}
\usepackage{subcaption}

\usepackage[accsupp]{axessibility}  

\usepackage[width=122mm,left=12mm,paperwidth=146mm,height=193mm,top=12mm,paperheight=217mm]{geometry}

\newcommand{\vz}{\bm{z}}
\newcommand{\vx}{\bm{x}}
\newcommand{\vy}{\bm{y}}
\newcommand{\vd}{\bm{d}}
\newcommand{\vu}{\bm{u}}
\newcommand{\vv}{\bm{v}}
\newcommand{\mX}{\bm{X}}
\newcommand{\mXrec}{\mX_\mathrm{rec}}
\newcommand{\mXfull}{\mX_\mathrm{full}}
\newcommand{\mV}{\bm{V}}
\newcommand{\mVpart}{\mV_\mathrm{part}}
\newcommand{\mVrec}{\mV_\mathrm{rec}}
\newcommand{\mVfull}{\mV_\mathrm{full}}
\newcommand{\mF}{\bm{F}}
\newcommand{\mFrec}{\mF_\mathrm{rec}}
\newcommand{\mFfull}{\mF_\mathrm{full}}
\newcommand{\mU}{\bm{U}}
\newcommand{\mN}{\bm{N}}
\newcommand{\mNpart}{\mN_\mathrm{part}}

\newcommand{\cG}{\mathcal{G}}
\newcommand{\cE}{\mathcal{E}}
\newcommand{\cD}{\mathcal{D}_{\mU}}

\begin{document}
\pagestyle{headings}
\mainmatter
\def\ECCVSubNumber{2523}  

\title{ Implicit Shape Completion via Adversarial Shape Priors} 

\titlerunning{Implicit Shape Completion via Adversarial Shape Priors}

\author{Abhishek Saroha \and
Marvin Eisenberger \and
Tarun Yenamandra \and
Daniel Cremers}
\authorrunning{Saroha et al.}

\institute{Technical University of Munich \\ Munich, Germany \\
\email{\{abhishek.saroha, tarun.yenamandra, cremers\}@tum.de}
\email{\{marvin.eisenberger\}@in.tum.de}
}

\maketitle

\begin{abstract}
We present a novel neural implicit shape method for partial point cloud completion. 
To that end, we combine a conditional Deep-SDF architecture with learned, adversarial shape priors. 
More specifically, our network converts partial inputs into a global latent code and then recovers the full geometry via an implicit, signed distance generator. Additionally, we train a PointNet++ discriminator that impels the generator to produce plausible, globally consistent reconstructions. In that way, we effectively decouple the challenges of predicting shapes that are both realistic, i.e. imitate the training set's pose distribution, and accurate in the sense that they replicate the partial input observations. In our experiments, we demonstrate state-of-the-art performance for completing partial shapes, considering both man-made objects (e.g. airplanes, chairs, ...) and deformable shape categories (human bodies). 
Finally, we show that our adversarial training approach leads to visually plausible reconstructions that are highly consistent in recovering missing parts of a given object.
\end{abstract}

\section{Introduction}

In recent times, neural implicit models experienced a steep increase in popularity as a tool for 3D shape modeling. 
The high flexibility and representation power of such approaches make them often preferable to other, alternative 3D representations like voxel grids, point clouds, polygonal meshes, or grid-based SDFs.
For once, neural implicit methods are grid-free and can, in theory, represent objects up to an arbitrary resolution. In practice, this enables them to adapt the level of precision to different regions of a given surface -- the resulting encoding is, at the same time, compact and precise.
In comparison to explicit surface representations like meshes, they can represent geometries of an arbitrary topology, only requiring the target surface to be watertight.
Moreover, implicit surfaces can be converted to other familiar representations in a straightforward manner, e.g. voxel grids by thresholding, or triangle meshes via the classical marching cubes algorithm~\cite{lorensen1987marchingcubes}.

A common approach in this line of work is to devise encoder-decoder architectures~\cite{mescheder2019occupancy,park2019deepsdf,peng2020convonet,chibane2020ifnet} to convert various types of input data into high-fidelity 3D surfaces. The decoder, in this context, predicts the occupancy probability or signed distance values for a set of xyz-point coordinates. This methodology has been used successfully to convert point clouds into watertight surfaces, reconstruct the 3D geometry from image observations or perform voxel super-resolution \cite{xu2019disn,chibane2020ifnet,saito2019pifu}. 
On the other hand, while such encoder-decoder approaches excel at reproducing objects from the training distribution, they often lack robustness when generalizing to unseen test poses. 

One of the key insights of our work is that we can extend the generalization capacity of neural implicit generative models by utilizing adversarial shape priors. Generative Adversarial Networks (GANs) \cite{goodfellow2014generative} are a ubiquitous tool for generative modeling on the 2D image domain. By and large, the popularity of GANs is based on their ability to generate novel data instances that imitate the style of the training distribution. While standard encoder-decoder architectures merely replicate the exact training samples, GANs are able to generate truly novel datapoints. Motivated by this observation, in this work we leverage the power of adversarial shape priors, in combination with a neural implicit architecture~\cite{park2019deepsdf,kleineberg2020shapegan}. 

A central open problem in the context of 3D generative models is shape completion from partial input observations. Common use-cases include shape reconstruction from range sensors like LIDAR, or completing partially occluded inputs. 
Many existing methods define shape completion as filling in local information, completing the backside of a front-facing human~\cite{chibane2020ifnet}, or completing halves of symmetric objects (e.g. left side of airplanes)~\cite{park2019deepsdf}. 
In contrast, we consider shape completion of point clouds that show truly arbitrary views of an object with varying degrees of partiality. By definition, this necessitates a holistic view on a given input observation: Completing a given snippet requires (1) encoding and categorizing the partial input point cloud (2) inferring the appropriate global geometry and (3) reconstructing the full shape, while taking both (1) and (2) into account. In an encoder-decoder architecture, (1)-(3) are strongly correlated. In our experiments, we demonstrate that, by including adversarial shape priors, our method is able to disentangle (1) and (2) to a certain extent. Overall, this leads to significantly more robust predictions on partial snippets of unseen test poses. There are a number of recent works that aim at increasing the fidelity of neural implicit surface decoders by leveraging more local information~\cite{peng2020convonet,chibane2020ifnet}. Our results indicate that this is not always preferable for shape completion, which requires modeling semantic information about a given object on the level of the global shape pose.

\subsubsection{Contribution} 
Our contributions can be summarized as follows:
\begin{enumerate}
\item We devise a novel neural network architecture based on implicit surfaces that predicts complete, watertight meshes from partial input point cloud observations.
\item We learn enhanced shape priors via generative adversarial learning. In that manner, our model effectively decouples the tasks of reconstructing meaningful shapes and fitting them to partial inputs.
\item Our method achieves state-of-the-art performance for completing partial point cloud observations of both man-made objects (lamps, chairs, \dots) and non-rigidly deformable shapes (human bodies).
\end{enumerate}

\section{Related work}

The field of 3D generative models experienced a surge in popularity over the last few years. In the following, we provide an overview of approaches most closely related to ours. We focus on related neural implicit models and methods that are specifically designed for 3D shape completion.

\subsubsection{Neural implicit models}
Implicit surface models have a long-standing history in computer vision, see~\cite{osher2003signed} for an overview of classical level-set approaches. Fuelled by the rise of deep learning, the pioneering works on occupancy networks~\cite{mescheder2019occupancy}, DISN~\cite{xu2019disn}  and Deep-SDF~\cite{park2019deepsdf} simultaneously proposed to represent implicit functions as a feed-forward neural network. This paradigm has a number of crucial advantages compared to other 3D representations. For once, deep implicit models are significantly more memory efficient than voxel grids~\cite{3dr2n2,hane2017hierarchical} or classical signed distance functions~\cite{osher2003signed}. In contrast to template-based~\cite{loper2015smpl,litany2018deformable,wang2018pixel2mesh,jiang2020shapeflow} and general mesh-based~\cite{groueix2018papieratlasnet} methods, they allow for reconstructing watertight surfaces without making any assumptions about the input object's topology.
While point cloud generation models have similar advantages~\cite{fan2017pointsetgen,yang2018foldingnet,achlioptas2018learningrepand3dpc}, they only produce sparse samples of a 3D surface without any notion of connectivity.

The initial works~\cite{mescheder2019occupancy,xu2019disn,park2019deepsdf} sparked countless follow-up works that build up on the idea of generating 3D shapes with neural implicits. There are a number of works combine neural implicit models with local, grid-based convolutions for different applications like creating human avatars~\cite{saito2019pifu}, representing 3D scenes~\cite{peng2020convonet} or standard object-level reconstruction~\cite{xu2019disn}. Similar to our approach,~\cite{kleineberg2020shapegan} learn an adversarial prior for 3D shape synthesis. \cite{niemeyer2019occupancy} reconstructs 4D shape sequences via learnable flow fields. \cite{yenamandra2021i3dmm} devise a morphable model for human faces based on neural implicits. 
Another related line of work~\cite{wang2021metaavatar,chen2021snarf,tiwari2021neural,saito2021scanimate} aims at reconstructing highly realistic human avatars by averaging scans in multiple poses via deformable human models like SMPL~\cite{loper2015smpl}. Note that, compared with such methods, our approach works out of the box for a broad range of object categories, since it makes no additional modeling assumptions that are exclusive to humans.

\subsubsection{Shape completion}
The central task we consider in this work is shape completion, i.e. reconstructing a full object from a partial observation. A standard approach to that end is learning shape priors via an encoder-decoder architecture and then predicting optimal reconstructions by fixing the learned decoder. This approach has been explored individually for both voxelized, man-made objects \cite{stutz2018learning} and deformable meshes with fixed topology~\cite{litany2018deformable}. Similarly,~\cite{park2019deepsdf} solve shape completion via a neural implicit encoder-decoder architecture. Our approach builds up on this methodology while additionally learning adversarial shape priors via a PointNet++ discriminator~\cite{qi2017pointnet++}. Another common approach is learning to generate full point clouds from partial ones via learnable decoders~\cite{yuan2018pcn,yu2021pointr} based on FoldingNet~\cite{yang2018foldingnet} or hierarchical shape decoding~\cite{tchapmi2019topnet}. Others propose to complete voxel grids via 3D convolutional networks~\cite{song2017semantic,dai2017_3depn}. Combining this idea with neural implicits,~\cite{chibane2020ifnet} learns hierarchical, voxel-based features which are subsequently decoded into an occupancy-based representation. 

\section{Method}

The overall objective of our method is to perform shape completion of partial point cloud observations. Specifically, the goal is to map a partial input point cloud $\mVpart\in\mathbb{R}^{n\times 3}$ to a full, reconstructed triangular mesh $\mXrec:=\bigl(\mVrec,\mFrec\bigr)$, where $\mVrec\in\mathbb{R}^{m\times 3}$ and $\mFrec\subset\mVrec\times\mVrec\times\mVrec$ are the sets of vertices and faces of $\mXrec$, respectively. We analogously write $\mXfull:=\bigl(\mVfull,\mFfull\bigr)$ for the full, watertight input training shapes. Furthermore, we denote $\mU=(\vu_i)_{i=1,\dots,k}\in\mathbb{R}^{k\times 3}$ as a set of randomly sampled, uniform points from the unit sphere $\vu_i\in\bigl\{\vx:\|\vx\|_2\leq 1\bigr\}$. During training, the points $\vu_i$ are used as query points to train the Deep-SDF signed distance prediction network. In the following, we define our neural network architecture in~\Cref{section:model_definition} and loss function in~\Cref{section:loss_function}. We further provide additional implementation details in~\Cref{section:implementation_details}.

\begin{figure}
    \centering
    \includegraphics[width=\linewidth]{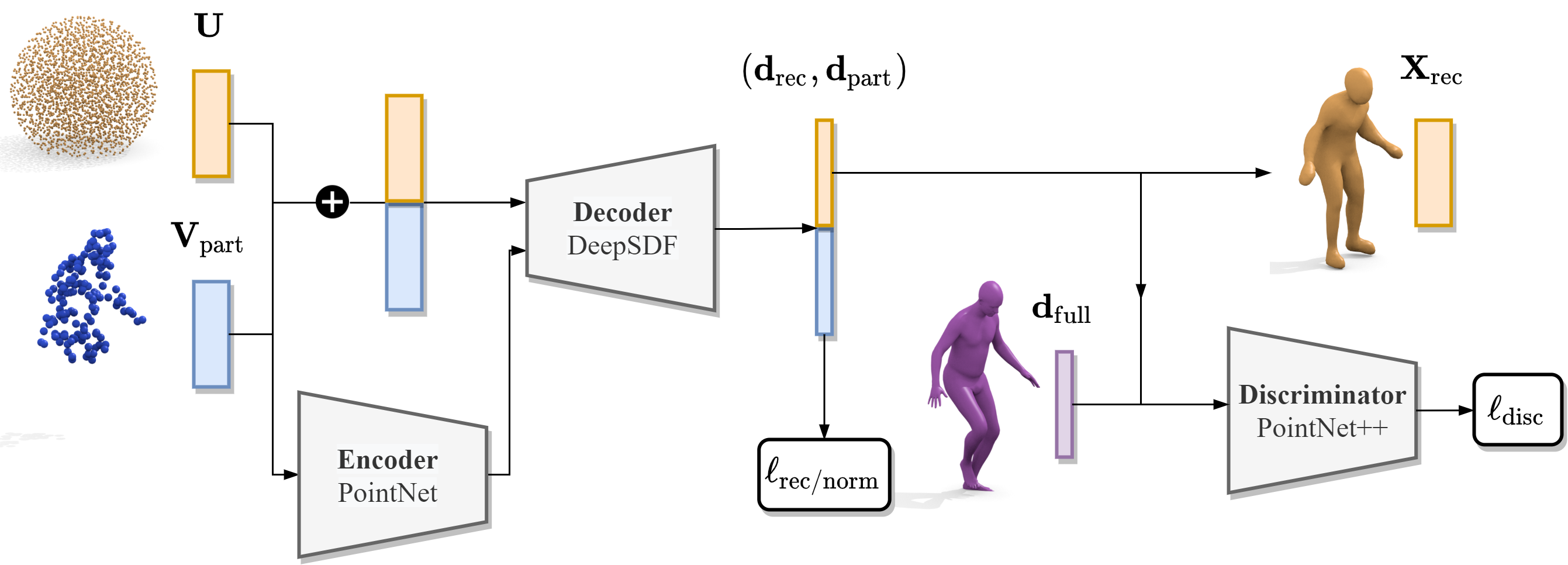}
    \caption{\textbf{Overview}. Our model takes a partial point cloud $\mVpart$ and generates signed distance predictions $\vd_{\mathrm{rec}}$ for a uniform set of points $\mU$, conditioned on the encoding $\vz:=\cE(\mVpart)$. Similarly, it predicts distances for the input points $\mVpart$ in order to generate a shape that matches the partial observations. At test time, the resulting distance field $\vd_{\mathrm{rec}}$ is converted to the full, reconstructed shape $\mXfull$.
    The discriminator incentivizes meaningful output shapes by learning to distinguish between generated distance fields $\vd_{\mathrm{rec}}$ and corresponding real samples $\vd_{\mathrm{full}}$.
    }
    \label{fig:model_overview}
\end{figure}

\subsection{Network}\label{section:model_definition}
The core of our method is an adversarial~\cite{goodfellow2014generative}, implicit surface neural network architecture~\cite{mescheder2019occupancy,park2019deepsdf}, see \Cref{fig:model_overview} for an overview. In the following, we define its individual components:

\subsubsection{Encoder}\label{model:encoder}
The encoder $\cE$ is responsible for converting the partial input point cloud $\mVpart$ into a high-dimensional latent code $\vz$. 

We can write this mapping $\cE$ as

\begin{equation}
\cE:\begin{cases}\mathbb{R}^{n\times 3}\to\mathbb{R}^{512}\\\mVpart\mapsto\vz,\end{cases}
\end{equation}

where $\mVpart \in \mathbb{R}^{n\times 3}$ is the partial input point cloud and $\vz \in \mathbb{R}^{{512}}$ is the $512$-dim predicted latent encoding. 
More specifically, we choose $\cE$ as a point cloud architecture based on the classification network of PointNet~\cite{qi2016pointnet}. In comparison to more sophisticated point cloud processing networks~\cite{qi2017pointnet++,wang2019dynamic,thomas2019kpconv}, PointNet has no local feature detection mechanism and mainly focuses on the global geometric structure of a considered object.
In fact, the main objective of the sub network $\cE$ is to encode the global pose context of $\mVpart$ whereas, detecting local features is of secondary importance. Additionally, this makes our method robust to varying input sampling densities, which we demonstrate in \Cref{fig:sampling_ablation}. Apart from being a straightforward choice, we therefore found empirically that the bias of PointNet towards global geometric features serves our purpose best in this context. 

\subsubsection{Generator}\label{model:generator}
The generator $\cG$ (or decoder) is responsible for generating 3D surfaces that match the underlying training distribution. The network architecture of $\cG$ is based on existing work on Deep-SDF decoders~\cite{kleineberg2020shapegan,park2019deepsdf}, see our supplementary material for more details. For a given set of query points $\vx_1,\dots,\vx_k$, which are provided to the network as the input, the generator $\cG$ predicts signed distance values to the target surface. 

In summary, we can define $\cG$ as

\begin{equation}\label{eq:generator}
      \cG:\begin{cases}\mathbb{R}^{3}\times\mathbb{R}^{512}\to\mathbb{R}\\(\vx_i, \vz) \mapsto d_i:=\mathrm{SDF}_{\cG,\vz}(\vx_i),\end{cases}
\end{equation}

where $\vx_1,\dots,\vx_k \in \mathbb{R}^{{k\times 3}}$ are the query points with the corresponding, predicted signed distance values $d_1,\dots,d_k \in \mathbb{R} $ and $\vz \in \mathbb{R}^{{512}}$ is a $512$-dim latent vector predicted by the encoder $\cE$. 

During training, we consider two different types of query points $\vx_i$ as inputs to $\cG$: uniformly sampled points $\vx_i:=\vu_i$ and partial input surface points $\vx_i:=\vv_{\mathrm{part},i}$. The former represents the predicted output shape $\mXrec$ and the latter is used to train the reconstruction loss, see \Cref{section:loss_function} for more details. 
By construction of \Cref{eq:generator}, the generator $\cG$ processes each input point $\vx_i$ independently.
For convenience, we therefore simply concatenate both of these signals $\mU=(\vu_i)_{i=1,\dots,k}\in\mathbb{R}^{k\times 3}$ and $\mVpart=(\vv_{\mathrm{part},i})_{i=1,\dots,n}\in\mathbb{R}^{n\times 3}$ in practice and obtain a single input point cloud $(\mU,\mVpart)\in\mathbb{R}^{(k+n)\times 3}$ with which we query $\cG$ during training. This corresponds to the orange/blue matrices in \Cref{fig:model_overview}. 

\subsubsection{Discriminator}
The final component of our method is the discriminator network $\cD$, playing the role of the critic that classifies generated point clouds into real/fake:

\begin{equation}
\cD:~\mathbb{R}^{k}\to\mathbb{R},~~\vd \mapsto c.
\end{equation}

The mapping $\cD$ takes a signed distance field $\vd=(d_1,\dots,d_k)\in\mathbb{R}^k$ as input. During training, we consider either the distances $\vd_{\mathrm{rec}}:=\cG(\mU,\vz)$ predicted by the generator or the ground-truth field $\vd_{\mathrm{full}}$ corresponding to the full training shapes $\mXfull$, see \Cref{section:loss_function} for more details. The subscript $\mU$ of the discriminator $\cD$ highlights its implicit dependency on the uniform query point cloud $\mU\in\mathbb{R}^{k\times 3}$. In practice, we concatenate $\mU$ and $\vd$ to a 4D input point cloud $(\mU,\vd)\in\mathbb{R}^{k\times 4}$ within the first layer of $\cD$.
The output score $c \in \mathbb{R}$ is used subsequently used to compute the GAN loss, see \Cref{section:loss_function} for more details. 

Following the common methodology of GANs on images, the $\cD$ acts as a critic that encourages $\cG$ to imitate the training shape distribution and synthesize more realistic samples. To that end, we chose the PointNet++ architecture~\cite{qi2017pointnet++} as the basis for the discriminator $\cD$. Compared to the standard PointNet network~\cite{qi2016pointnet}, PointNet++ decomposes a given input point cloud into multiple subregions in a hierarchical manner. Hence, it is much more effective at detecting local features, which is a major prerequisite for obtaining realistic reconstructions. Note, that this is in stark contrast to the encoder network $\cE$, whose main objective is detecting the global pose features of a considered object. To further illustrate this point, we show a comparison of the two different discriminators, in an ablation study, see our supplementary material.

\subsection{Loss function}\label{section:loss_function}

To train our model, we define the following loss function:

\begin{equation}\label{eq:tot_loss}
\ell\bigl(\cE,\cG,\cD\bigr) :=\ell_{\mathrm{GAN}}\bigl(\cG,\cD\bigr)+\lambda_{\mathrm{rec}}\ell_{\mathrm{rec}}\bigl(\cE,\cG\bigr)+\lambda_{\mathrm{norm}}\ell_{\mathrm{norm}}\bigl(\cE,\cG\bigr).
\end{equation}

Motivated by the standard GAN methodology, our generator $\cG$ and discriminator $\cD$ are trained in an alternating manner, playing the role of each other's adversary. The discriminator tries to distinguish samples generated by the generator from real training samples, which in turn drives $\cG$ to synthesize more realistic shapes.

\begin{equation}\label{eq:minmax}
    \min_{\cE,\cG}\max_{\cD}\ell\bigl(\cE,\cG,\cD\bigr).
\end{equation}

In practice, we optimize \Cref{eq:minmax} in an alternating manner, taking single update steps of $\cE,\cG$ and $\cD$ for a single batch, respectively. In the remainder of this section, we define the individual components referred to in \Cref{eq:tot_loss}.

\subsubsection{GAN loss}

We devise the GAN loss component of our method by following the popular image model Wasserstein-GAN (WGAN)~\cite{arjovsky2017wgan}

\begin{equation}\label{eq:gen_loss}
\ell_{\mathrm{GAN}} = {\mathbb{E}_{\vd_{\mathrm{full}}}}\bigl[\cD (\vd_{\mathrm{full}})\bigr]-{\mathbb{E}_{\mVpart}}\bigl[\cD \bigl(\cG(\mU,\cE(\mVpart))\bigr)\bigr]+\lambda_{\mathrm{GP}}GP
\end{equation}

where we sample from both the training distributions of real samples $\vd_{\mathrm{full}}$ and partial conditional inputs $\mVpart$. Here, $GP$ denotes a gradient penalty term, defined as 

\begin{equation}\label{eq:grad_penality}
\mathbb{E}_{\lambda\sim\mathcal{U}(0,1)}\biggl[\bigl(\bigl\|\nabla_{\vd} \cD \bigl((1-\lambda)\vd_{\mathrm{full}}+\lambda\vd_{\mathrm{rec}}\bigr)\bigr\|_{2}-1\bigr)^{2}\biggr],
\end{equation}

where $\mathcal{U}(0,1)$ denote the uniform distribution on the interval $(0,1)$. Furthermore, $\vd_{\mathrm{full}}$ and  $\vd_{\mathrm{rec}}=\cG(\mU,\cE(\mVpart))$ are defined as the signed distances of the full input training shapes and the distance field predicted by the generator, respectively. Main idea of this gradient penalty term is to enforce the approximate Lipschitz property $\nabla_{\vd} \cD(\cdots)\approx 1$ on the predictions of the $\cD$ as a soft version of gradient clipping, see~\cite{arjovsky2017wgan} for further details. 

\subsubsection{Reconstruction loss}
In addition to the adversarial loss, we employ a reconstruction loss on the outputs of the generator. 

\begin{equation}\label{eq:rec_loss}
    \ell_{\mathrm{rec}} = {\mathbb{E}_{\mVpart}}\bigl[\bigl\|\cG\bigl(\mVpart, \cE(\mVpart)\bigr)\bigr\|_2\bigr],
\end{equation}
 
where $\mVpart$ are the surface points of the partial input point cloud. The purpose of this loss is to enforce the predicted signed distances of points on the input surface $\mVpart$ to be as close as possible to zero. 

Note, that $\ell_\mathrm{GAN}$ prevents the generator to opt for the degenerate solution $\cG(\cdot,\cdot)\equiv 0$ and instead take a sample from the learned prior distribution, for which the points $\mVpart$ are as close as possible to the surface of the predicted shape. 

\subsubsection{Normal loss}
We further utilize the outer normals of the partial input point clouds $\mVpart$ 
\begin{equation}\label{eq:norm_loss}
\ell_{\mathrm{norm}} = 
{\mathbb{E}_{\mVpart}}\bigl[\bigl\|\nabla_{\mV}\cG\bigl(\mVpart, \cE(\mVpart)\bigr)-\mNpart\bigr\|_2\bigr],
\end{equation}

where $\mNpart \in \mathbb{R}^{(n)\times 3}$ denotes the surface normals corresponding to the point set $\mVpart$. As demonstrated in~\cite{gropp2020igr}, the loss in \Cref{eq:norm_loss} encourages the generator $\cG$ to replicate small scale features present in the training shape distribution. In practice, we extract $\mNpart$ from the full training mesh before sampling the partial point cloud $\mVpart$.

\subsection{Implementation details}\label{section:implementation_details}

\subsubsection{Training protocol}\label{section:training_protocol}
To train our model for the task of shape completion, we consider a training set, defined as a collection of full, watertight triangular meshes $\bigl\{\mXfull^{(1)},\dots,\mXfull^{(N)}\bigr\}$, where $\mXfull^{(i)}=\bigr(\mVfull^{(i)},\mFfull^{(i)}\bigr)$, see the purple shape in \Cref{fig:model_overview}.
In each training iteration, we sample a random half-space that removes parts of the full input shape to generate the partial inputs $\mVpart$, see the blue point cloud in \Cref{fig:model_overview}. The relative ratio of retained points after cut-off is chosen randomly from the uniform distribution $\mathcal{U}(0.5,1)$. In that way, the model learns to complete shapes with varying degrees of partialities. At test time, we consider cut-off ratios sampled from $\mathcal{U}(0.5,0.55)$ to focus on the challenging case of $\approx 50\%$ retained points. We show additional partiality settings at test time in our ablation study, see \Cref{table:extreme_partiality}.

Our model is trained in a coarse-to-fine manner: We start training the network with $n=m=1024$ sampled points for both $\mU$ and $\mVpart$. After every 300 epochs, we double the number of points in $\mU, \mVpart$ to $n=m=2048,4096,\dots,32768$ points. For refinement, we append another 1000 epochs on the highest resolution $n=m=32768$. This progressive training schedule is motivated by previous works~\cite{karras2018progressive,kleineberg2020shapegan} which show that it improves the training stability and overall synthesis quality. 

\subsubsection{Data preparation}
As specified in \Cref{sec:experiments}, we consider datasets that focus on both man-made objects and deformable human bodies, respectively.
For data preparation, we follow the protocol described by~\cite{kleineberg2020shapegan,park2019deepsdf}. The input to our method is a training set of triangular meshes $\bigl\{\mXfull^{(1)},\dots,\mXfull^{(N)}\bigr\}$. We first scale the vertices of the meshes $\mXfull^{(i)}$ to a unit sphere for data normalization. For each training shape, we then uniformly sample a set of $100,000$ points $\vu_j$ in the ambient domain and compute the ground-truth signed distance values $\vd_{\mathrm{full},j}$ of each point $\vu_j$ from the surface boundary of $\mXfull^{(i)}$. To convert the mesh $\mXfull^{(i)}$ into a point cloud, each mesh is rendered from multiple camera angles to obtain depth buffers, which are projected back into the object space, see~\cite{kleineberg2020shapegan} for more details. 
We then sample the partial inputs $\mVpart$, as well as their corresponding normals $\mNpart$, as described in the previous paragraph.
Shapes which had either less than 1\% of the samples points inside the shape, or a discontinuous SDF, are discarded from the respective training and test sets. 

In summary, for every processed mesh we get two input point clouds. the first is the point cloud $\mU$, signified by the orange box in \cref{fig:model_overview}, which contains randomly sampled points around the input object. $\mVpart$ and $\mNpart$, on the other hand, are the surface points and corresponding outer normals sampled from the meshes $\mXfull$. 

\subsubsection{Parameter specifications}
During training, we use RMSProp~\cite{rmsprop} as the optimizer for updating the parameters of the various modules of our model. For the discriminator optimization, the optimizer has a learning rate of 1e-5, while for the encoder and the generator, the learning rate was set to 1e-3. We also set the relative weight of the reconstruction loss to $\ell_{\mathrm{rec}}=8e-3$, while the weight of the normal loss component is set to be $\ell_{\mathrm{norm}}=0.01$. The Gradient Penalty defined \eqref{eq:grad_penality} is weighed by a factor of $\lambda_{\mathrm{GP}}=10$.

\section{Experiments}\label{sec:experiments}

\subsection{Datasets}\label{sec:datasets}

We choose two dataset to asssess our method's ability to complete shape from both rigid, man made object classes and non-rigidly deformable shape categories. Specifically, we consider the following datasets: 

\subsubsection{ShapeNet}
The ShapeNet dataset~\cite{shapenet2015} is a large collection of man-made 3D models. Its objects are classified into various different categories, such as airplanes, sofas, chairs, cars, etc. Beyond the objects' geometries, the ShapeNet dataset also provides annotations such as correspondences, keypoints, various orientation vectors, and more. More specifically, we use ShapeNetCore (v1), which is a subset of the ShapeNet dataset, containing approximately 51,300 models that span 55 different categories. For our method, we specifically focus on three different classes from the full dataset, which were lamps, benches, and displays.

\subsubsection{SURREAL Dataset}
SURREAL~\cite{varol17_surreal} is a large-scale dataset which contains about 6 million 3D models of synthetic humans. In addition to the pose information, they also provide details such as optical flow, depth, normals, segmentation of body parts etc. For our purposes, we use a subset of $1886$ shapes for training and $536$ different poses as our test set.

\subsection{Evaluation}
\subsubsection{Metric}

We evaluate different methods by measuring the Chamfer Distance (CD) between the ground truth models $\mXfull$ and the predicted shapes $\mXrec$. The CD measures the distance between two point clouds, defined as the mean distance of each individual point on the first point cloud to the second point set:

\begin{equation}\label{eq:chamfer_distance}
\mathrm{CD}\left(\mV_{1}, \mV_{2}\right)=\frac{1}{\left|\mV_{1}\right|} \sum_{\vx \in \mV_{1}} \min _{\vy \in \mV_{2}}\|\vx-\vy\|_{2}^{2}+\frac{1}{\left|\mV_{2}\right|} \sum_{\vy \in \mV_{2}} \min _{\vx \in \mV_{1}}\|\vx-\vy\|_{2}^{2}.
\end{equation}

\subsubsection{Baselines}\label{sec:baselines}
We compare our approach to a broad range of recent approaches that allow for shape completion. Specifically, we show state-of-the-art methods for three relevant categories of baselines methods:
\begin{itemize}
    \item \textbf{Occupancy Networks~\cite{mescheder2019occupancy}} ONet is a classical encoder-decoder method that like ours encodes a partial input point cloud into a global feature vector $\vz$ and subsequently predicts a full geometry via a neural implicit decoder.
    \item \textbf{Convolutional Occupancy Networks~\cite{peng2020convonet}} The second baseline ConvONet is based on ONet, but additionally leverages local features refinement on a latent 3D feature grid with learnable 3D convolutions.
    \item \textbf{PCN~\cite{yuan2018pcn}} On top of showing state-of-the-art neural implicit approaches, we also compare our methods to a recent point cloud based shape completion approach to provide a more complete picture.
\end{itemize}

\subsubsection{Discussion}
\begin{figure}
    \centering
    \includegraphics[width=0.48\linewidth]{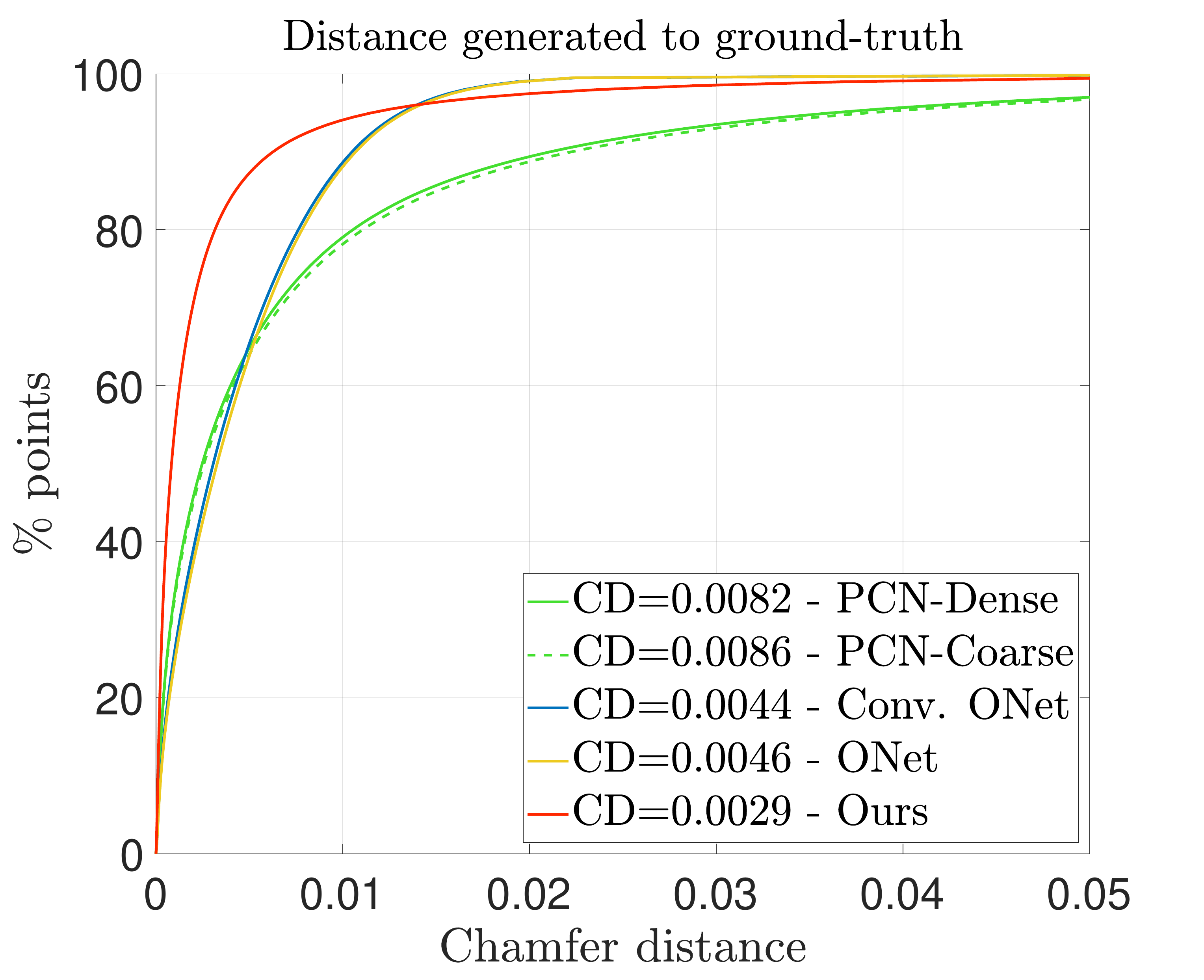}
    \includegraphics[width=0.48\linewidth]{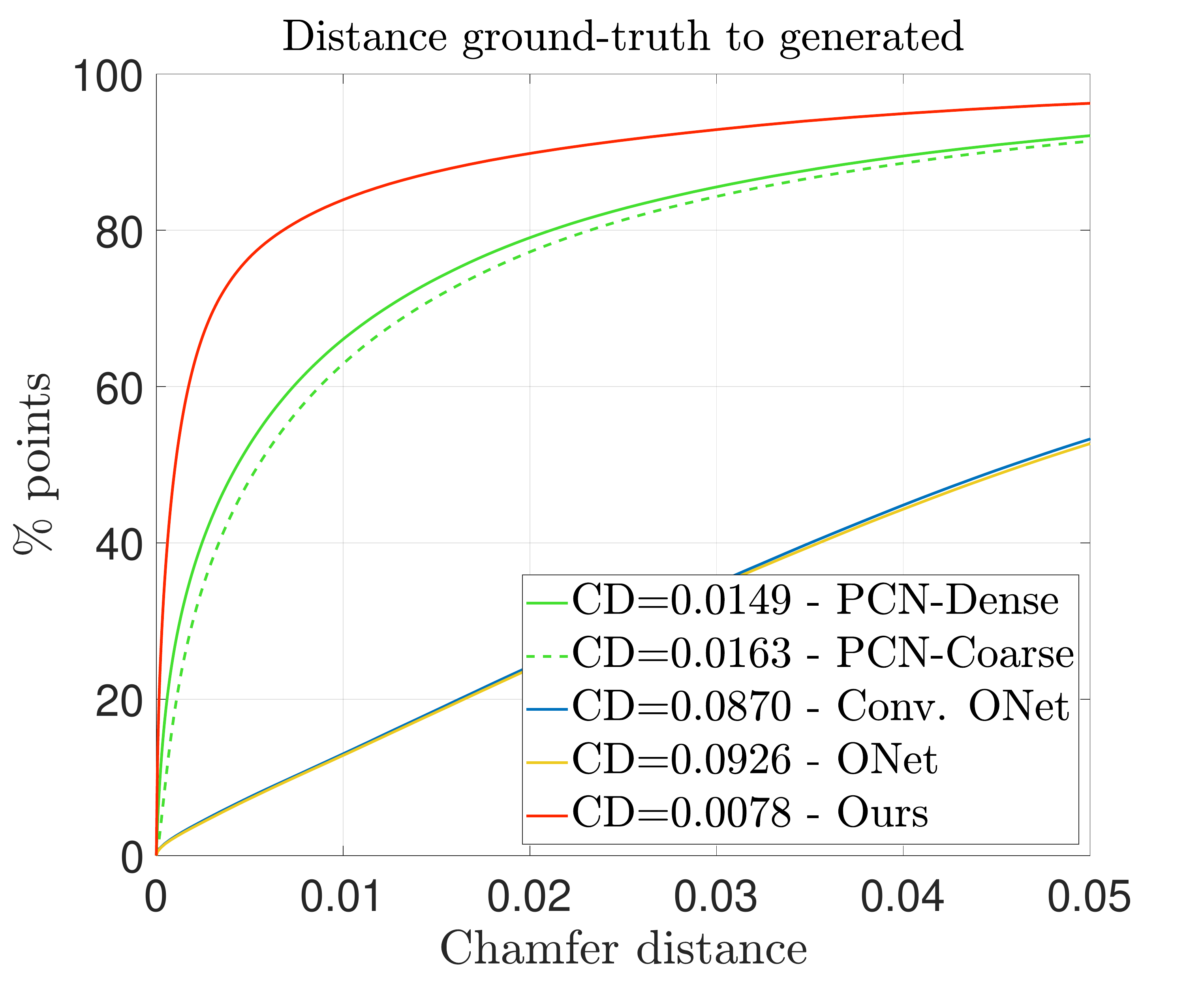}
    \caption{We show cumulative Chamfer distance curves on SURREAL for our method, compared to the baselines ONet~\cite{mescheder2019occupancy}, ConvONet~\cite{peng2020convonet} and PCN~\cite{yuan2018pcn}. Each curve shows the ratio of points below a certain CD threshold for CD values between $[0,0.05]$. The left plot reports the distance of the generated shape $\mXrec$ to the input $\mXfull$, corresponding to the first term in \Cref{eq:chamfer_distance}. Vice-versa, the right plot shows the distance of $\mXfull$ to $\mXrec$. Analogous to \Cref{table:chamfer_dist}, we further report the mean CD values in the lower right corner (legend) of each plot. 
    }
    \label{fig:surreal_cumulative}
\end{figure}

\begin{figure}
\vspace{40pt}
\begin{center}
\begin{overpic}
[width=\linewidth]{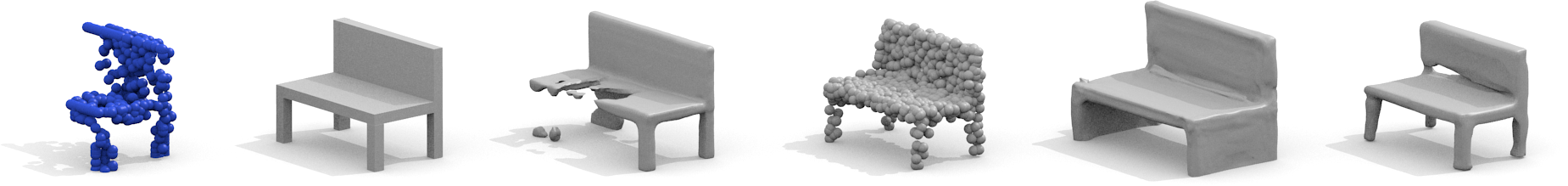}
\put(3,16){$\mVpart$}
\put(20,16){$\mXfull$}
\put(35,16){ConvONet}
\put(56,16){PCN}
\put(73,16){ONet}
\put(90,16){Ours}
\end{overpic}
\\[10pt]
\includegraphics[width=\linewidth]{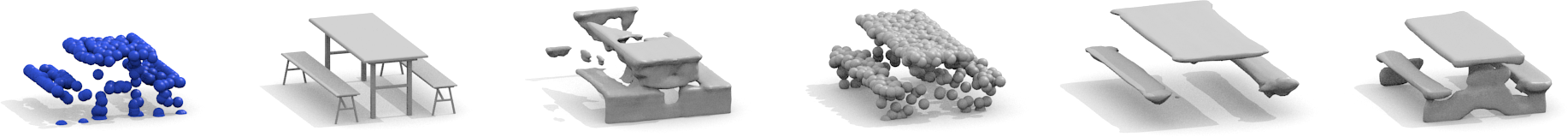}
\\[30pt]
\includegraphics[width=\linewidth]{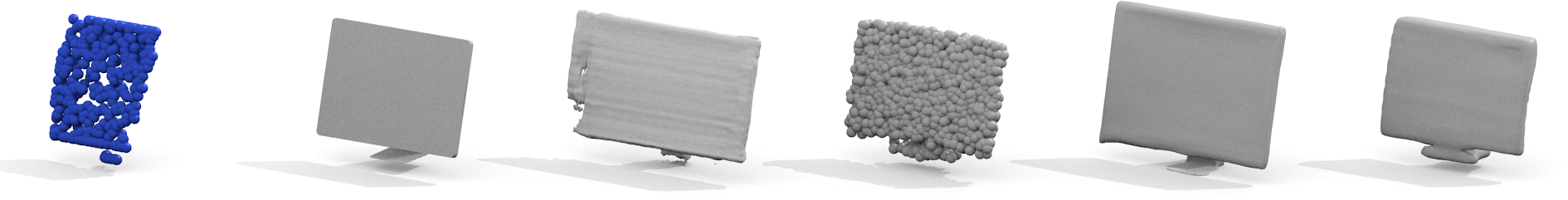}
\\[10pt]
\includegraphics[width=\linewidth]{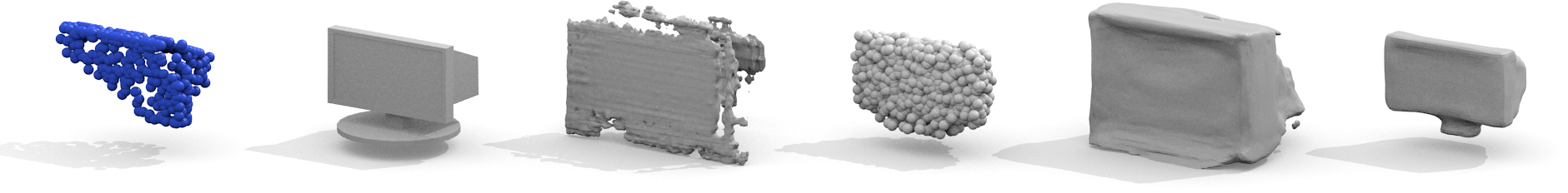}
\\[30pt]
\includegraphics[width=\linewidth]{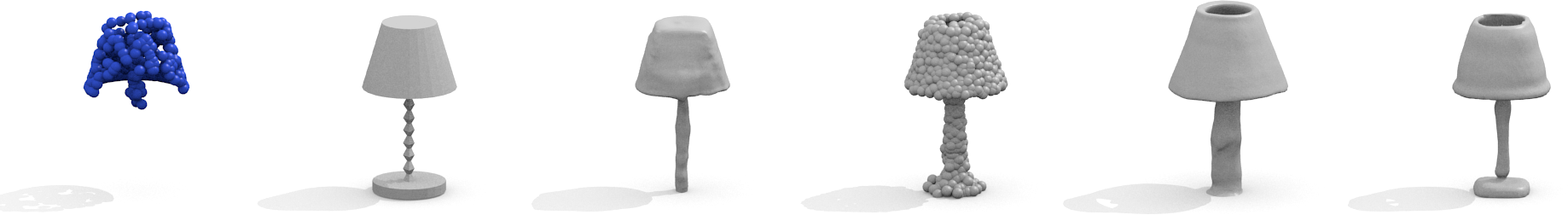}
\\[10pt]
\includegraphics[width=\linewidth]{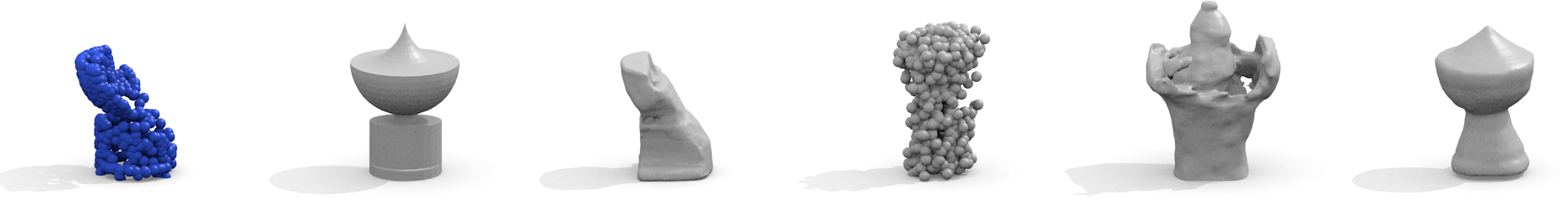}
\end{center}
\caption{\textbf{Qualitative comparison -- ShapeNet.} We show two separate qualitative examples for all three object classes of ShapeNet considered in \Cref{table:chamfer_dist}, namely 'Bench', 'Display', Lamp'. These results highlight clearly that, compared to other existing methods, our learned adversarial shape priors allow for a more robust generalization performance on unseen test samples. 
}
\label{fig:qual_shapenet}
\end{figure}

\begin{figure}
\begin{center}
\begin{overpic}
[width=\linewidth]{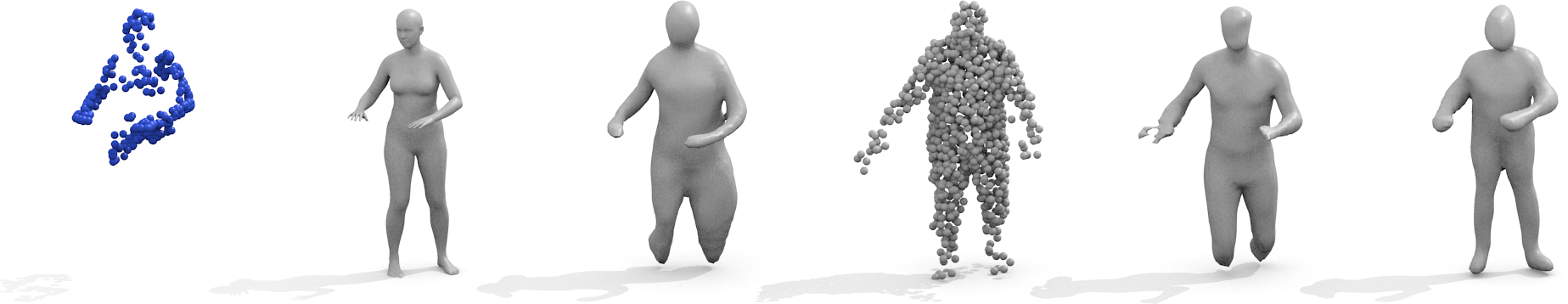}
\put(6,23){$\mVpart$}
\put(23,23){$\mXfull$}
\put(38,23){ConvONet}
\put(59,23){PCN}
\put(76,23){ONet}
\put(93,23){Ours}
\end{overpic}
\includegraphics[width=\linewidth]{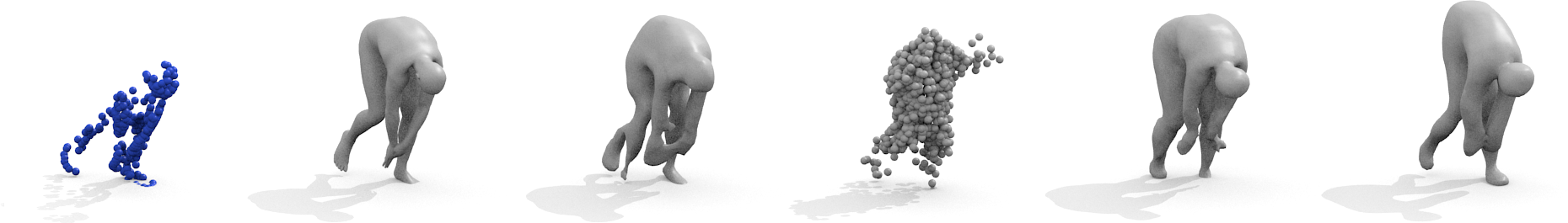}
\includegraphics[width=\linewidth]{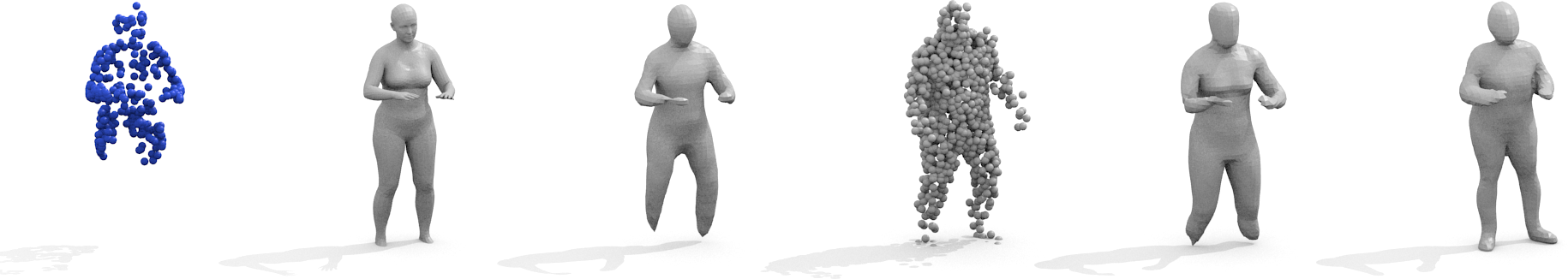}
\includegraphics[width=\linewidth]{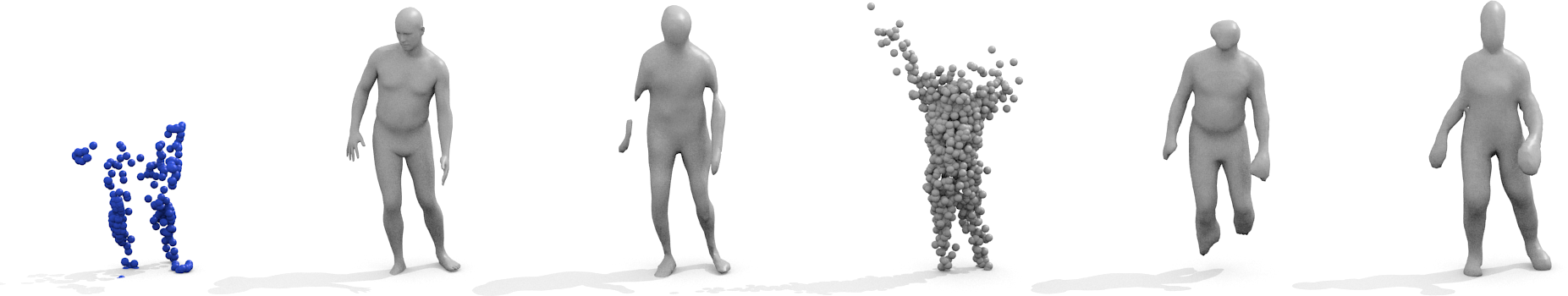}
\includegraphics[width=\linewidth]{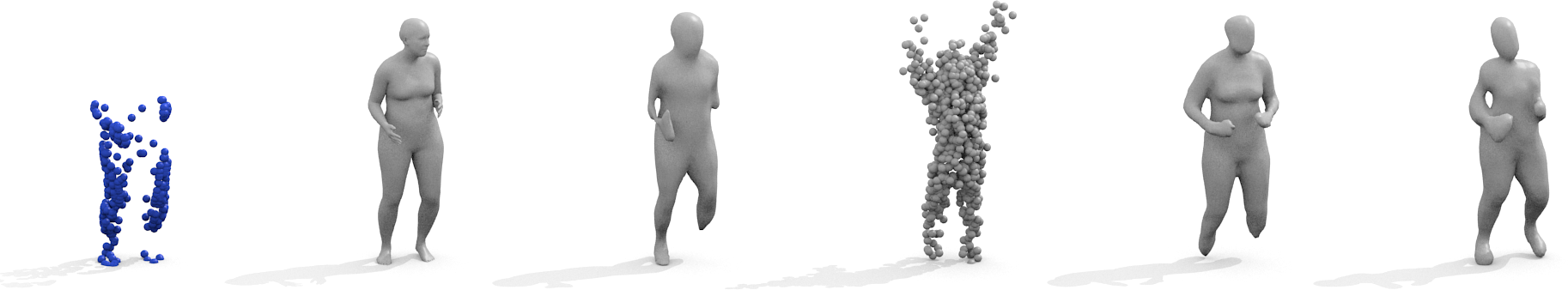}
\includegraphics[width=\linewidth]{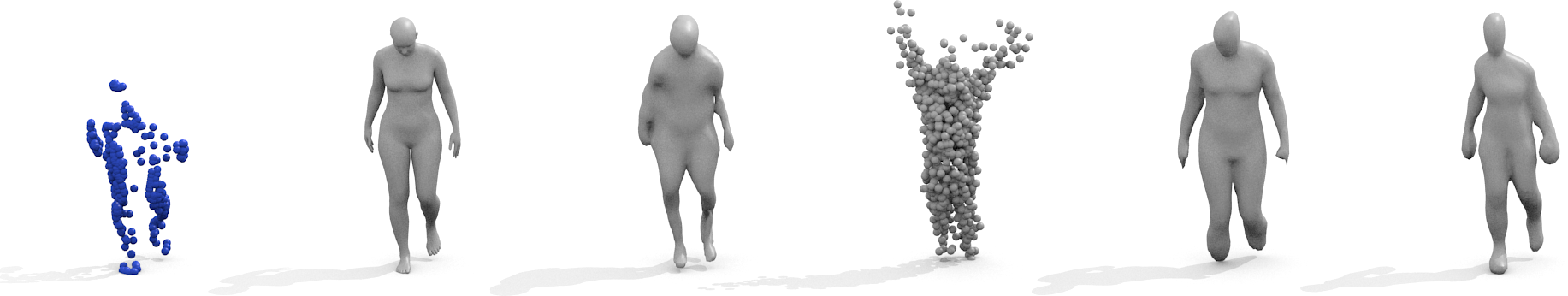}
\includegraphics[width=\linewidth]{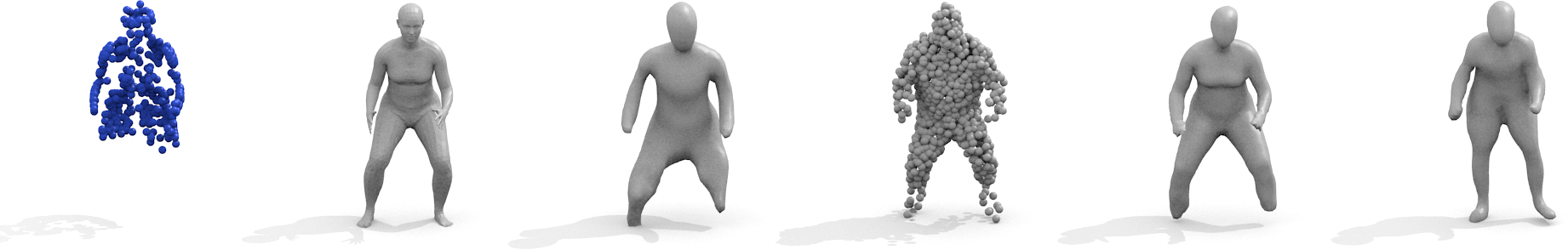}
\end{center}
\caption{\textbf{Qualitative comparison -- SURREAL.} We show several different qualitative completion results corresponding to the analysis in \Cref{fig:surreal_cumulative}. We observe that, while the considered baselines are able to predict a meaningful global pose for most partial inputs, the predictions from our method are generally more robust. Overall, our approach tends to produce more stable results at the extremities (arms and legs).
}
\label{fig:qual_surreal}
\end{figure}

We show a quantitative comparison between our method and the considered baselines on ShapeNet in \Cref{table:chamfer_dist}. We further provide corresponding qualitative comparisons in \Cref{fig:qual_shapenet}. Analogously, we show results on SURREAL in \Cref{fig:surreal_cumulative}. To give a more complete picture, in the latter case we also plot the total cumulative CD curves for each method, additionally to showing the mean CD values. Qualitative comparisons on SURREAL are shown in \Cref{fig:qual_surreal}.

As seen in \Cref{table:chamfer_dist} and \Cref{fig:surreal_cumulative}, our method achieves excellent quantitative performance on rigid objects such as Shapenet as well as non-rigid humans of the SURREAL dataset. Especially on the deformable human shapes from SURREAL, our method outperforms the considered baselines by a significant margin of $91\%$ and $58\%$, respectively. Aside from the quantitative performance, our qualitative comparisons in \Cref{fig:qual_shapenet} and \Cref{fig:qual_surreal} further highlight the superior visual quality of our obtained reconstructions on both datasets. 

On SURREAL, the baselines ONet\cite{mescheder2019occupancy} and ConvONet\cite{peng2020convonet} frequently fail to complete specific missing details, such as the feet and hands, see \Cref{fig:qual_surreal}. 
On the man-made objects from ShapeNet, our method shows a much more robust capacity to generalize to unseen test poses. We believe that this can, by and large, be attributed to our learned adversarial shape priors. While encoder-decoder based baselines like ONet simply learn to encode the full training pose $\mXfull$, they often fail to generalize to previously unseen object geometries and instead opt for reconstructing the closest example from the training set. While ConvONet is, in principle, able to reproduce fine scale details, in our observation it also proves to be less stable when faced with partial input geometries -- it is mainly designed for full object level reconstruction or reconstructing large-scale scenes. The local 3D convolution refinement proves to be a disadvantage for shape completion since, while providing more expressivity, it also tends to make the decoder less robust for imperfect inputs.

\begin{table*}
\centering
\resizebox{0.80\linewidth}{!}{
\begin{tabular}{l|lllll}
\toprule[0.1em]
CD ($\downarrow$) & Ours ~~~~ & ONet~~~~  & ConvONet & PCN-Coarse & PCN-Dense\\
\hline
\hline
Lamp           & 9.49  & 9.99     & 25.92 & 9.20 & \textbf{8.73}     \\
Display           & 16.51  & \textbf{15.79} & 17.22 & 18.47  & 18.45    \\
Benches       & \textbf{18.63} & 18.77 & 21.00 & 22.13 & 21.81   \\
\bottomrule[0.1em]
\end{tabular}
}
\vspace{10pt}
\caption{A comparison of our method with other baselines approaches for shape completion on the Shapenet dataset. For each setting, we show the mean CD, averaged over 30000 points. The results in the Table are scaled by a factor of $10^3$ for readability. The partial inputs $\mVpart$ consist of $n=500$ points for Ours, ONet\cite{mescheder2019occupancy} and ConvONet\cite{peng2020convonet}. For PCN, we show results for their two separate network settings, see~\cite{yuan2018pcn} for more details. Additionally, to maintain consistency with the training of PCN, we test this specific baseline on a resolution of $n=8000$ points on the same partial snippets as the other methods.}
\label{table:chamfer_dist}
\end{table*}

\subsection{Ablation study}\label{section:ablation}

To provide a deeper understanding our how our approach learns to complete partial shapes, we show two additional experimental settings. Moreover, we show additional ablation studies in our supplementary material.

\subsubsection{Sampling density}
\begin{figure}
    \centering
    \includegraphics[width=0.8\linewidth]{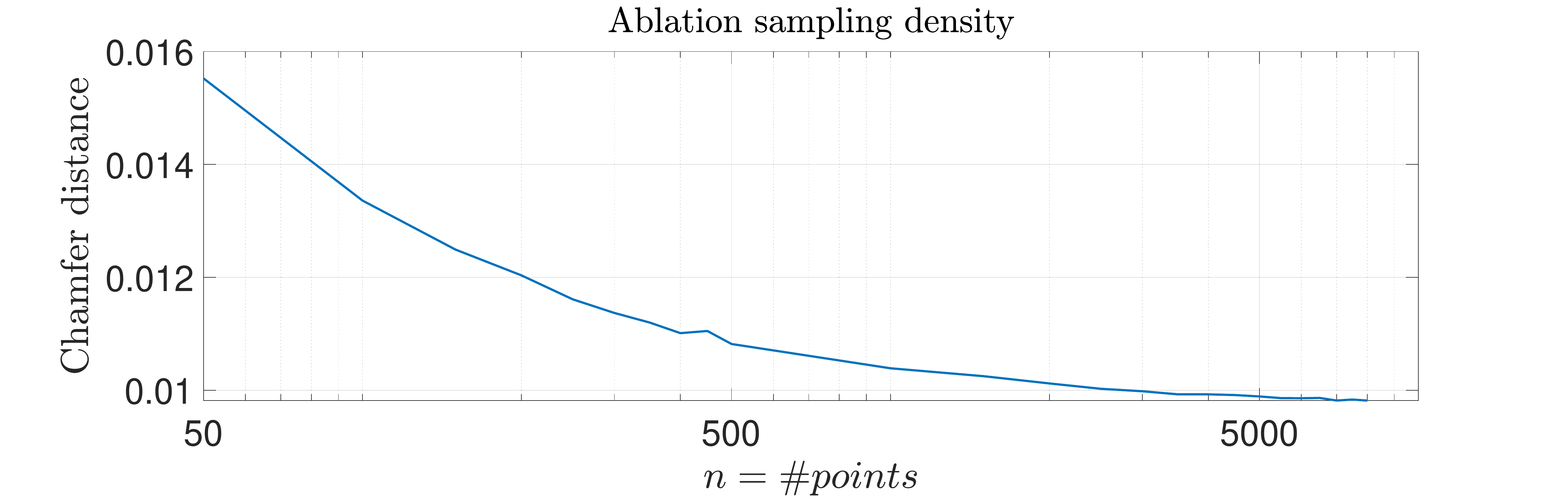}
    \vspace{-5pt}
    \caption{We show the effect of varying the density of sampled points $n$ (x-axis) on the partial inputs $\mVpart$. For each $n$, we denote the test set performance of our model in terms of the CD on SURREAL, analogous to the results in \Cref{fig:surreal_cumulative}. These results show that even for extremely sparse samplings $n=50$ the CD to the ground-truth meshes does not change by a significant margin. For $n>1000$, the performance plateaus, indicating that increasing the resolution beyond this point does not benefit our method.
    }
    \label{fig:sampling_ablation}
\end{figure}

We assess how varying the number of test points $n$ of the partial inputs $\mVpart$ affects our performance on SURREAL. More specifically, we gradually increase the number of input points from $n=50$ to $n=8000$ and report the resulting accuracies in \Cref{fig:sampling_ablation}. The overall insight here is that our method is fairly robust to sparser inputs. We attribute this to two factors: For once, our model is trained in a progressive manner which allows it to learn the underlying adversarial priors at various scales, see \Cref{section:implementation_details} for more details. Moreover, by choosing a PointNet architecture as our encoder $\cE$, the network has a certain degree of robustness built-in. PointNet is based on independent feature refinement on the input points and global feature pooling ($\max$ pool), both of which are relatively stable under varying point densities.

\subsubsection{Partial cut-off ratio}
The second ablation we consider investigates how the degree of partiality affects the prediction results at test time. More specifically, we take our model trained on SURREAL with partiality ratios of $\mathcal{U}(0.5,1)$ and test it for different degrees of partial test shapes $\mVpart$ see \Cref{table:extreme_partiality}. 

\begin{table*}
\centering
\resizebox{0.85\linewidth}{!}{
\begin{tabular}{l|lllllllllll}
\toprule[0.1em]
Partial \%        & 0.05   & 0.1    & 0.2   & 0.3   & 0.4   & 0.5  & 0.6  & 0.7  & 0.8  & 0.9  & 1.0  \\ \hline \hline
CD & 253.883 &
170.93 &
58.08 &
32.42 &
20.93 &
13.19 &
8.24 &
5.65 &
4.24 &
3.32 &
2.88 \\ \hline
Gen to g.t. CD & 6.54   & 6.03   & 6.48  & 5.62  & 4.46  & 3.36 & 2.47 & 2.01 & 1.78 & 1.62 & 1.62 \\ \hline
G.t. to gen CD & 247.33 & 164.89 & 51.59 & 26.80 & 16.46 & 9.82 & 5.76 & 3.63 & 2.45 & 1.70 & 1.26 \\

\bottomrule[0.1em]
\end{tabular}
}
\vspace{5pt}
\caption{Ablation study on SURREAL for test shapes with varying degrees of partialites in the range $[0.05,1]$. In each setting, we train our model on the standard training set with ratios $[0.5,1]$ of retained points, see \Cref{section:implementation_details} for more details.}
\label{table:extreme_partiality}
\end{table*}

\section{Conclusion}

We presented a novel approach to shape completion that combines a neural implicit surface architecture with learnable, adversarial shape priors. In that manner, our network effectively decouples the task of shape completion into (1) reconstructing objects that are meaningful and (2) fitting shapes from the learned distribution to a conditional, partial observation. We demonstrate state-of-the-art performance on both ShapeNet and SURREAL, highlighting our network's flexibility of completing both rigid, man-made objects and non-rigidly deformable shape categories. Our method yields robust predictions for various degrees of partiality and different sampling densities. Finally, we show a variety of qualitative comparisons that highlight the superior visual quality of the shape prior learned by our adversarial model.

\section*{Acknowledgements}
This work was supported by the ERC Advanced Grant SIMULACRON and the Munich School of Machine Learning.

\clearpage

\bibliographystyle{splncs04}
\bibliography{ms}

\clearpage

\appendix
\section{Network architecture details}

\subsubsection{Encoder}
The encoder $\cE$ converts an input point cloud $\mVpart\in\mathbb{R}^{n\times 3}$ with $n$ points into a high-dimensional latent encoding $\vz\in\mathbb{R}^{512}$.
The specific architecture is based on PointNet~\cite{qi2016pointnet}. It comprises 4 linear layers followed by leaky ReLU non-linearities and a max-pool operation over the points $n$ after the last linear layer. The linear layers change the feature dimension from 3, representing the spatial coordinates, to 64, 128, 256, and finally to 512 dimensions. The max pool operation after the last linear layer reduces the entire feature tensor to a single latent vector $\vz\in\mathbb{R}^{512}$ per input shape.

\subsubsection{Generator}
The generator $\cG$ is responsible for predicting the SDF values for a set of query points $\vx_1,\dots,\vx_k \in \mathbb{R}^{{k\times 3}}$. Apart from the points $\vx_i$, the second input to $\cG$ is a vector $\vz\in\mathbb{R}^{512}$, containing a latent representation of the partial input point cloud $\mVpart$ obtained from the encoder $\cE$. $\cG$ comprises 8 linear layers, all operating on 128 dimension features, where each linear layer is followed by layer normalization~\cite{ba2016layernorm} and non-linear activation. In each layer, we use leaky ReLU as the non-linearity, since these were shown to increase the generation robustness in the context of generative adversarial networks. It furthermore contains a linear layer that maps the input latent code of the partial shape $\vz$ from a 512 (output size of the encoder defined in \ref{model:encoder}) to a 128 dimensional vector. Similarly, the points $\vx_i$ are converted from $3$ input dimensions to $128$ latent dimensions in a linear layer. The two 128-dim latent vectors produced by the initial input layers are added up before being passed to the second layer.

To facilitate gradient flow and increase training stability, we use skip connections~\cite{he2015resnet} between the input and the 5th layer. Similar to the first layer, $\vz$ is first passed through a linear layer before being added to the output of the fourth layer. 

\subsubsection{Discriminator}
We use a standard PointNet++~\cite{qi2017pointnet++} classification architecture without batch normalization layers, as our discriminator $\cD$. The module is composed of what the authors call 'set abstraction layers', where the points are recursively sampled, grouped and passed through individual PointNet layers. The purpose of this scheme is to learn the features in a hierarchical way, which allows it to extract fine grained details. After a series of abstraction layers and subsequent pooling, the resulting feature tensor is passed through three linear layers, the final one of which outputs a scalar value corresponding to the discriminator's output classification of true/false.

\section{Results on additional ShapeNet classes}
In \Cref{table:chamfer_dist} (main paper), we show a quantitative comparison between our approach and various baselines. The experiment in the main paper comprises of 3 specific classes of ShapeNet that we deemed to be most representative and most suitable for capturing the pose and geometry diversity in the whole dataset. For completeness, we also show quantitative results on the remaining 9 classes of ShapeNet here, see \Cref{table:extra_shapenet_quant}. The training set of each class is defined by randomly selecting 500 samples, while the evaluation was performed on the full test sets.

\begin{table}[]
\vspace{-15pt}
\centering
\scalebox{0.95}{
\begin{tabular}{l|lllll}
\toprule[0.1em]
CD ($\downarrow$) & Ours ~~~~ & ONet~~~~  & ConvONet & PCN-Coarse & PCN-Dense\\
\hline
\hline
Chair          & \textbf{10.49} & 15.59   & 12.43           & 11.33    & 10.75 \\
Cabinet          & \textbf{18.27}   & 24.09 & 20.03           & 21.92   & 21.34 \\
Table            & \textbf{19.38}            & 23.13   & 19.52 & 22.05   & 21.23 \\
Car              & 2.34          & 2.32   & 3.08           & 1.79 & \textbf{1.41} \\
Vessel           & 5.01         & 4.63    & 4.45           & 4.11  & \textbf{3.71}   \\
Loudspeaker      & \textbf{17.42}  & 24.84 & 19.70          & 19.23  & 18.59    \\
Sofa             & \textbf{11.42}  & 11.93   & 11.65         & 12.49   & 11.90  \\
Telephone        & 13.68        & 15.72 & \textbf{12.61} & 14.40 & 14.32        \\
Airplane        & 2.49         &  2.32 & 2.42 & 2.16 & \textbf{1.88} \\
\bottomrule[0.1em]
\end{tabular}}
\vspace{5pt}
\caption{We show additional quantitative comparisons of our method to our baselines ONet~\cite{mescheder2019occupancy}, ConvONet~\cite{peng2020convonet}, and PCN~\cite{yuan2018pcn} for the task of shape completion on the 9 ShapeNet classes in addition to the 3 classes in \Cref{table:chamfer_dist}. Similar to \Cref{table:chamfer_dist}, the average CD is scaled by a factor of $10^3$ for readability. We observe that different classes have results on different CD scales, varying by up to one order of magnitude. Aside from the different complexities of individual categories, this can be in part attributed to varying vertex densities in the ground-truth meshes.}
\label{table:extra_shapenet_quant}
\end{table}

\section{Additional ablations}

We perform a number of additional ablations to provide further insights and assess the role of the various sub-components of our network.

\subsection{Training set size}
\begin{figure}
    \vspace{-5pt}
    \centering
    \includegraphics[width=0.72\linewidth]{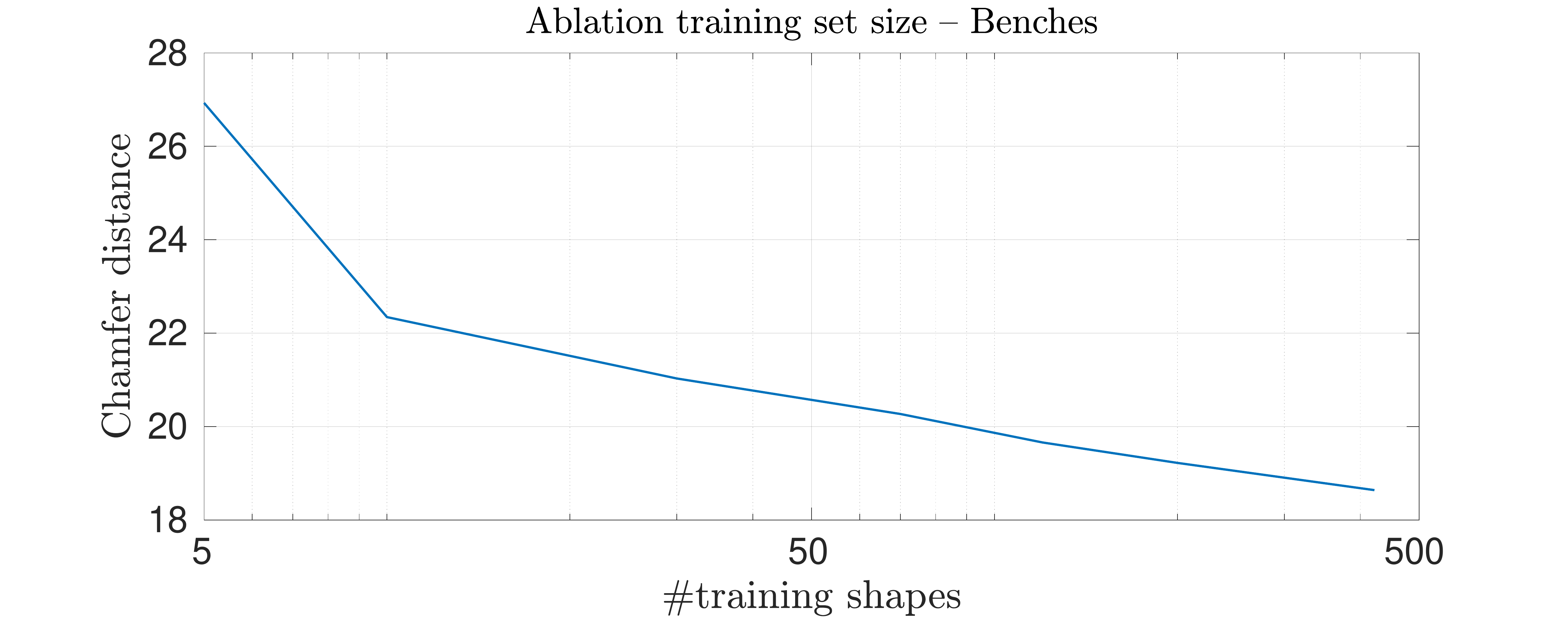}
    \vspace{-5pt}
    \caption{An ablation study assessing how the number of training shapes impacts the quantitative performance of our model. Specifically, we report the test set CD for the task of shape completion on the 'Benches' class of ShapeNet. The setting of $422$ training shapes (right end of the curve) is equivalent to the result reported in \Cref{table:chamfer_dist}, with a CD of $18.63$. We conclude that our method is relatively robust, even if the number of samples in the training set are reduced considerably.
    }
    \label{fig:training_data_ablation}
\end{figure}

We perform an ablation to study the behaviour of our network in the low-training data regime, see \Cref{fig:training_data_ablation}. Our results indicate that decreasing the number of training shapes does not significantly impact the performance of our model. In fact, while the CD error increases for small $\#$ training samples, the training remains relatively stable even when up to $\approx99\%$ of the training data is removed.

\subsection{Network ablation}
Finally, we study how two specific components of our network impact the reconstruction accuracy. For once, we compare how replacing our discriminator $\cD$ based on PointNet++~\cite{qi2017pointnet++} with a more primitive PointNet~\cite{qi2016pointnet} discriminator changes the results. For both discriminators $\cD$, we assess the impact of removing the normal loss $\ell_{\mathrm{norm}}$ from our total loss function $\ell$. 
 
\begin{figure}
\centering
\scalebox{0.9}{
\begin{subfigure}{0.4\linewidth}
\begin{tabular}{l|l}
\toprule[0.1em]
Setting & CD ($\downarrow$) \\ 
\hline
(i) PN w/o $\ell_{\mathrm{norm}}$ & 12.30 \\  (ii) {PN w/ $\ell_{\mathrm{norm}}$} & 12.90 \\ (iii) PN++ w/o $\ell_{\mathrm{norm}}$& 10.82 \\ (iv) Ours & 10.70 \\
\bottomrule[0.1em]
\end{tabular}

\label{table:disc_ablation_surreal}
\end{subfigure}
\begin{subfigure}{0.58\linewidth}
    \begin{overpic}
    [width=\linewidth]{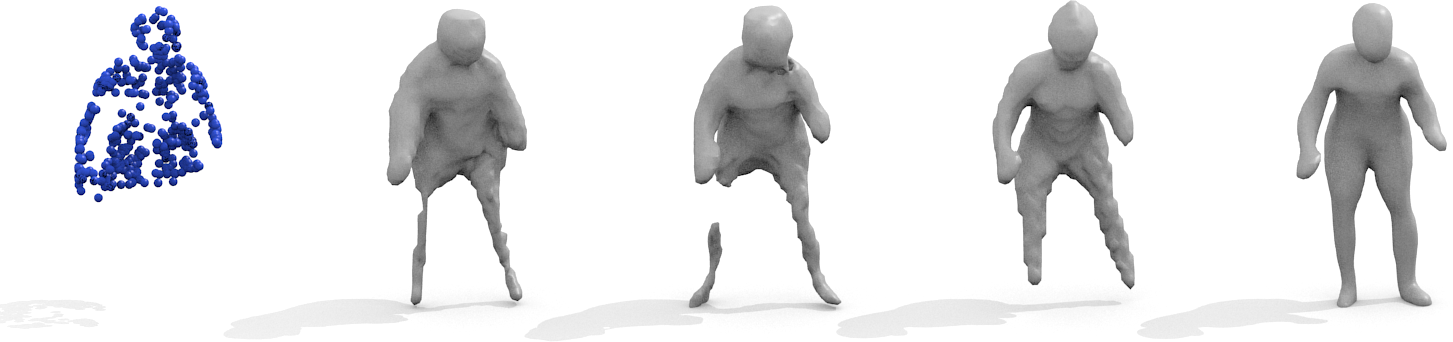}
    \put(6,29){$\mVpart$}
    \put(30,29){(i)}
    \put(50,29){(ii)}
    \put(70,29){(iii)}
    \put(90,29){(iv)}
    \end{overpic}
   
\end{subfigure}}
\caption{We show a quantitative and qualitative ablation study to assess the role of two design choices in our model. We assess the impact of replacing the PointNet++~\cite{qi2017pointnet++} discriminator architecture $\cD$ with the more primitive PointNet~\cite{qi2016pointnet}. Additionally, we show a version of both networks where we remove the normal loss component $\ell_{\mathrm{norm}}$. Overall, the four settings are: (i) PointNet without $\ell_{\mathrm{norm}}$, (ii) PointNet with $\ell_{\mathrm{norm}}$, (iii) PointNet++ without $\ell_{\mathrm{norm}}$, and (iv) PointNet++ with $\ell_{\mathrm{norm}}$. The latter case (iv) is equivalent to our full model, as introduced in the main paper. We show a quantitative comparison on the test set of SURREAL (left side), as well as a qualitative sample reconstruction (right side). These results indicate that both the PointNet++ discriminator and the normal loss are crucial for an optimal shape completion performance. 
}
\label{figure:disc_ablation_overview}
\end{figure}

Results are summarized in \Cref{figure:disc_ablation_overview}. We report the resulting test set performance, equivalent to the shape completion experiment on SURREAL in \Cref{fig:surreal_cumulative} in the main paper. Additionally, we show sample reconstructions to assess the qualitative impact of the different settings.
It is immediately evident that there is a noticeable advantage of `Ours' which is based on PointNet++ and includes the normal loss $\ell_{\mathrm{norm}}$. The PointNet discriminator fails to capture fine details of the reconstructed shapes and is relatively ineffective at inpainting missing parts.
On the other hand, without the normal loss $\ell_{\mathrm{norm}}$, the reconstructions lack fine details, especially at salient regions such as the feet and hands of the human. These trends are also clearly reflected in the quantitative comparisons in \Cref{figure:disc_ablation_overview}. 

\end{document}